\documentclass[AMA,Times1COL]{WileyNJDv5} 

\articletype{Article Type}%

\received{Date Month Year}
\revised{Date Month Year}
\accepted{Date Month Year}
\journal{Journal}
\volume{00}
\copyyear{2025}
\startpage{1}

\begin{document}

\title{DeepMpMRI: Tensor-decomposition Regularized Learning for Fast and High-Fidelity Multi-Parametric Microstructural MR Imaging}







\author[1]{Wenxin Fan}
\author[2]{Jian Cheng}
\author[3]{Qiyuan Tian}
\author[1]{Ruoyou Wu}
\author[4]{Juan Zou}
\author[5]{Weixin Si}
\author[6]{Zan Chen}
\author[1]{Shanshan Wang}

\authormark{Wenxin \textsc{et al.}}
\titlemark{DeepMpMRI: Tensor-decomposition Regularized Learning for Fast and High-Fidelity Multi-Parametric Microstructural MR Imaging}

\address[1]{\orgdiv{Paul C. Lauterbur Research Center for Biomedical Imaging}, \orgname{Shenzhen Institutes of Advanced Technology, Chinese Academy of Sciences}, \orgaddress{\state{Shenzhen}, \country{China}}}


\address[2]{\orgdiv{State Key Laboratory of Complex \& Critical Software Environment,}, \orgname{Beihang University}, \orgaddress{\state{Beijing}, \country{China}}}

\address[3]{\orgdiv{School of Biomedical Engineering}, \orgname{Tsinghua University}, \orgaddress{\state{Beijing}, \country{China}}}

\address[4]{\orgdiv{School of Physics and Electronic Science}, \orgname{Changsha University of Science and Technology}, \orgaddress{\state{Changsha}, \country{China}}}

\address[5]{\orgdiv{Faculty of Computer Science and Control Engineering}, \orgname{Shenzhen University of Advanced Technology}, \orgaddress{\state{Shenzhen}, \country{China}}}

\address[6]{\orgdiv{Department of Information Engineering}, \orgname{Zhejiang University of Technology}, \orgaddress{\state{Hangzhou}, \country{China}}}

\corres{Shanshan Wang, Paul C. Lauterbur Research Center for Biomedical Imaging, Shenzhen Institutes of Advanced Technology, Chinese Academy of Sciences, Shenzhen 518055, China. 
\email{sophiasswang@hotmail.com}}



\abstract[Abstract]{Deep learning has emerged as a promising approach for learning the nonlinear mapping between diffusion-weighted MR images and tissue parameters, which enables automatic and deep understanding of the brain microstructures. However, the efficiency and accuracy in estimating multiple microstructural parameters derived from multiple diffusion models are still limited since previous studies tend to estimate parameter maps from distinct models with isolated signal modeling and dense sampling. This paper proposes DeepMpMRI, an efficient framework for fast and high-fidelity multiple microstructural parameter estimation from multiple models using highly sparse sampled q-space data. DeepMpMRI is equipped with a newly designed tensor-decomposition-based regularizer to effectively capture fine details by exploiting the high-dimensional correlation across microstructural parameters. In addition, we introduce a Nesterov-based adaptive learning algorithm that optimizes the regularization parameter dynamically to enhance the performance. DeepMpMRI is an extendable framework capable of incorporating flexible network architecture. Experimental results {on the HCP dataset and the Alzheimer's disease dataset} both demonstrate the superiority of our approach over 5 state-of-the-art methods in simultaneously estimating multi-model microstructural parameter maps for DKI and NODDI model with fine-grained details both quantitatively and qualitatively, achieving 4.5 - 15 $\times$ acceleration compared to the dense sampling of a total of 270 diffusion gradients.}

\keywords{diffusion MRI, microstructure estimation, tensor-SVD, adaptive learning}

\jnlcitation{\cname{%
\author{Taylor M.},
\author{Lauritzen P},
\author{Erath C}, and
\author{Mittal R}}.
\ctitle{On simplifying ‘incremental remap’-based transport schemes.} \cjournal{\it J Comput Phys.} \cvol{2021;00(00):1--18}.}

\maketitle

\renewcommand\thefootnote{}
\footnotetext{\textbf{Abbreviations:} dMRI, diffusion magnetic resonance imaging; DTI, diffusion tensor imaging; DKI, diffusion kurtosis imaging; NODDI, Neurite Orientation Dispersion and Density Imaging; AK, axial kurtosis; RK, radial kurtosis; MK, mean kurtosis; KFA, kurtosis fractional anisotropy; OD, rientation dispersion; $V_{ic}$, intra-cellular volume fraction; $V_{iso}$, isotropic / cerebrospinal fluid volume fraction.}

\renewcommand\thefootnote{\fnsymbol{footnote}}
\setcounter{footnote}{1}

\section{Introduction}\label{sec:introduction}

{Diffusion} MRI (dMRI) plays a vital role in characterizing in vivo brain microstructures non-invasively. Diffusion-weighted images (DWIs) are sensitive to the random displacement of water molecules within a voxel, probing tissue on scales significantly lower than image resolution \cite{kiselev2017fundamentals}. The diffusion MR signal in a voxel is considered a composite of signals from various compartments. By carefully designing signal models that relate tissue microstructure to diffusion signals, the organization of the neuronal tissue can be inferred from the observed measurements by model fitting \cite{alexander2019imaging}. The capability to detect subtle changes in brain tissue is particularly crucial in establishing biomarkers for the early-stage diagnosis of neurodegenerative diseases \cite{camacho2024exploiting, soskic2023dti}. 

{Diffusion tensor imaging (DTI) \cite{basser1994mr} is a well-established technique to quantify and analyze diffusion MRI signals by implicitly assuming that water molecule diffusion follows a Gaussian distribution of diffusion displacement, which oversimplifies the diffusion process. To better take into account the complexity of the white matter, more advanced representations or models were introduced to better take into account the complexity of the white matter. 
Diffusion kurtosis imaging (DKI) \cite{jensen2005diffusional} extends DTI naturally by including a kurtosis tensor to characterize the non-Gaussian properties of water diffusion, such as the presence of diffusion restrictions, reflected in the DKI-derived metrics including axial kurtosis (AK), radial kurtosis (RK), mean kurtosis (MK), and kurtosis fractional anisotropy (KFA). These derived parameters are shown to be more sensitive to microstructural alterations in diseases and have shown promising results in clinical applications \cite{pang2024dki, wang2011parkinson}. 
Besides DKI, several biophysical models have been proposed to achieve more specific and biologically interpretable characterizations of tissue microstructure, including the Intravoxel Incoherent Motion (IVIM) model \cite{le1986mr}, AxCaliber model \cite{assaf2008axcaliber}, the Spherical Mean Technique (SMT) model \cite{kaden2016quantitative}, and the Neurite Orientation Dispersion and Density Imaging (NODDI) model \cite{zhang2012noddi}. Specifically, the NODDI model distinguishes three microstructural environments: intra-neurite, extra-neurite, and cerebrospinal fluid compartments, thus enhancing the sensitivity to changes in the brain and providing specific markers that reflect tissue complexity and heterogeneity. 

Due to the complicated model design, advanced signal representations or models usually require prolonged imaging protocols with a large number of diffusion samples and multiple b-values. The reliable estimation of tissue microstructure described by these models may require close to or more than 100 diffusion gradients with multiple b-values, which requires a very long acquisition time (about an hour for 270 diffusion gradients) \cite{zhang2012noddi, kaden2016multi, kaden2016quantitative}. 
Thus, in clinical settings where imaging time is often limited to just a few minutes (about 30 diffusion gradients or less)\cite{li2024deep}, accomplishing accurate estimation of tissue microstructure with a reduced number of diffusion samples can be challenging.
Moreover, the attenuation of the dMRI signal makes the diffusion data suffer from low signal-to-noise ratio (SNR) issues, especially in high-b-value imaging \cite{le2006artifacts}, resulting in further degradation of the accuracy of subsequent quantitative imaging and analysis. Fitting advanced diffusion models with noisy and fewer measurements is generally a non-convex optimization problem, potentially having several local minima of the objective function \cite{novikov2018modeling}. Jelescu et al. \cite{jelescu2016degeneracy} evidenced that the estimation is an ill-posed inverse problem for clinically feasible dMRI acquisitions.

Recent studies have highlighted the value of integrating microstructural parameters derived from multiple diffusion models \cite{boonsuth2023feasibility, palacios2020evolution}, as enabling a more comprehensive characterization of tissue composition and microstructural organization. In this study, we combine DKI and NODDI to jointly probe brain microstructure. Although both models rely on the same underlying anatomical structures and water diffusion properties, they provide complementary insights: DKI captures the overall diffusion anisotropy by modeling non-Gaussian properties, whereas NODDI offers compartment-specific estimation across three distinctive microstructural environments. The synergy between these models facilitates a more detailed and biologically informative assessment of tissue complexity. As presented in Fig. \ref{correlation}, despite being derived from different modeling assumptions, the metrics reflect shared microstructural substrates, particularly axonal integrity and myelination, highlighting the main biological interdependence and complementarity.

However, the added benefit of multi-parameter microstructure imaging from multiple diffusion modeling is so far largely unexplored.
Deep learning (DL) has shown potential in microstructure estimation with sparse measurements \cite{golkov2016q, gibbons2019simultaneous, fandeephibrid, park2021diffnet, hashemizadehkolowri2022jointly, ye2019deep, ye2020improved, tian2020deepdti,li2024dimond, li2021superdti, chen2020estimating, chen2022hybrid, chen2023deep, yang2023towards,zheng2023microstructure, li2024deep}. The q-space deep learning (q\_DL) \cite{golkov2016q} directly maps dMRI signals to microstructural parameters using a limited number of q-space samples through a multilayer perceptron (MLP). Note that Golkov et al.\cite{golkov2016q} suggested using a separate MLP  for each microstructure parameter. The convolutional neural networks (CNNs) were also utilized to predict high-quality scalar diffusion metrics using a small amount of diffusion data \cite{tian2020deepdti,li2021superdti,li2024dimond}. Beyond the data-driven mapping approaches, model-based neural networks incorporating domain knowledge have been introduced to improve network performance and interpretability. Specifically, the model-based neural network is designed by unfolding the conventional optimization process of a mathematical model to construct deep networks. Ye et al. \cite{ye2019deep, ye2020improved} proposed patch-wise dictionary-based DL approaches to estimate parametric diffusion maps of NODDI and SMT separately. Chen et al. \cite{chen2020estimating, chen2022hybrid, yang2023towards} used a subset of q-space to estimate the NODDI parameters by explicitly considering the q-space geometric structure. Despite the promising achievements, the efficiency and accuracy of estimating multiple microstructural parameters derived from various diffusion models remain limited, as previous studies have primarily focused on estimating each model's metrics individually with isolated signal modeling. Although some research has employed multiple diffusion models learning \cite{gibbons2019simultaneous, fandeephibrid, park2021diffnet, hashemizadehkolowri2022jointly}, they tend to directly output multi-microstructural parameters without considering the correlation among these high-dimensional parameters.

\begin{figure}[!t]
\centerline{\includegraphics[width=0.5\textwidth]{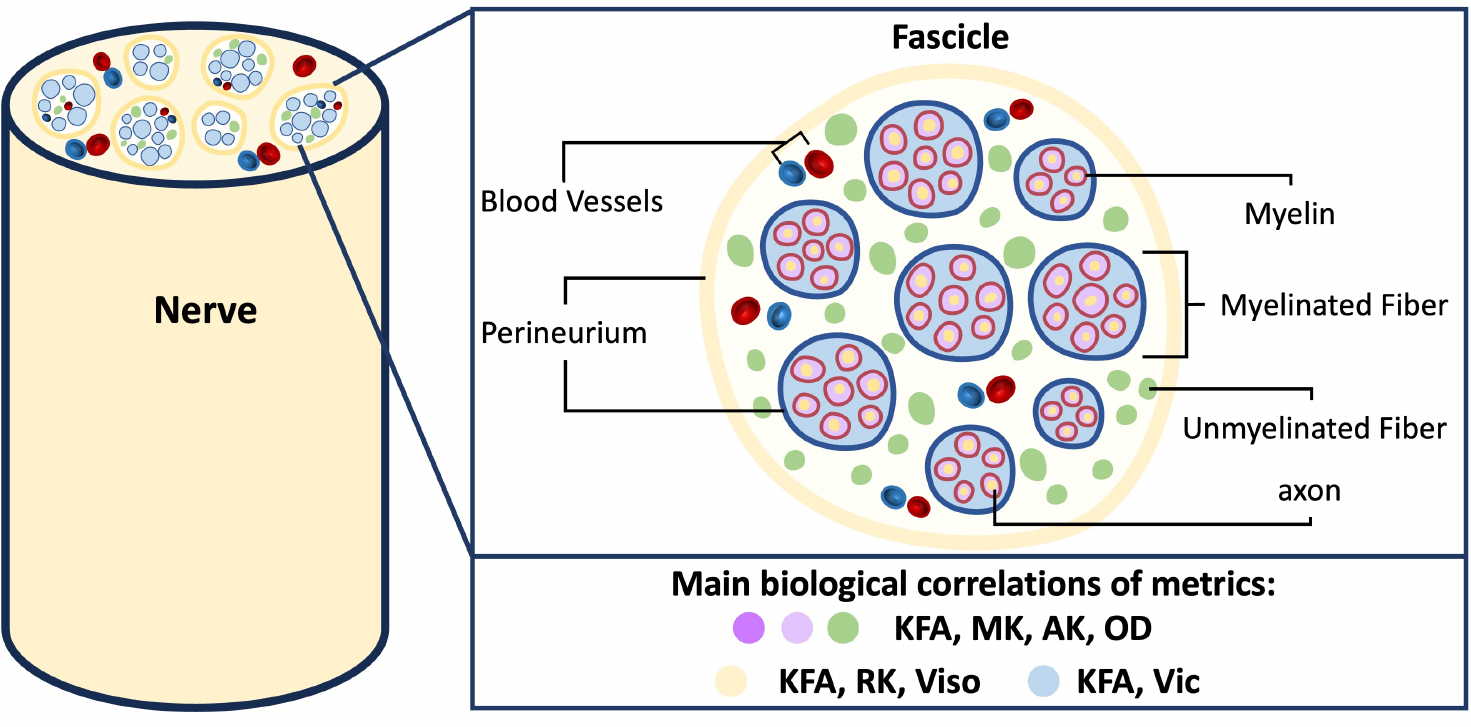}}
\caption{Schematic illustration of DKI and NODDI-derived microstructural parameter correlation in the nerve cross-sectional anatomy. 
}
\label{correlation}
\end{figure} 

Motivated by these, we propose a deep learning-based framework for diffusion MRI named DeepMpMRI, which simultaneously estimates multiple microstructural parameters derived from various diffusion models. DeepMpMRI incorporates a newly designed tensor-decomposition-based regularization (TDR) to exploit the spatial coherence and correlation among multiple parameters derived from different models. TDR preserves the inherent high-dimensional structure of multi-parameter tensors and utilizes tensor-SVD \cite{kolda2001orthogonal, kolda2009tensor, chen2009tensor} to uncover the relationship between them, enhancing the performance of microstructure estimation. Additionally, the regularization hyperparameter should be properly selected since it greatly impacts network performance. Therefore, we propose a lightweight adaptive hyperparameter learning algorithm integrating the Nesterov-based gradient method to streamline the optimization process in our task.


Overall, the main contributions of our work are as follows:
\begin{itemize}
\item We design a novel \textbf{t}ensor-\textbf{d}ecomposition-based \textbf{r}egularization (TDR) to exploit underlying high-order correlations from different dimensions in multiple microstructural parameters derived from various diffusion models. Targeting the alignment between predicted parameters and reference in tensor singular values effectively captures fine-grained details while suppressing noise.
\item We propose a \textbf{N}esterov-based \textbf{a}daptive \textbf{l}earning \textbf{a}lgorithm (NALA) that optimizes the regularization parameter dynamically, enabling more efficient hyperparameter tuning and better performance. Our method is efficient in terms of both memory and complexity and thus can be easily integrated into existing DL frameworks.
\item The proposed DL-based framework DeepMpMRI facilitates high-fidelity multi-model microstructural parameter estimation for various diffusion models using sparsely sampled q-space data. DeepMpMRI is a highly extendable framework that can accommodate diverse diffusion models and utilize a flexible network architecture as the backbone. In the following parts, we demonstrate the effectiveness of our method by estimating DKI-derived parameters and NODDI-derived parameters as an example. Experiments indicate that our method outperforms 5 state-of-the-art methods both quantitatively and qualitatively. 
\end{itemize}

The rest of the paper is organized as follows. Section \ref{sec:related_work} reviewed relevant literature. Section \ref{sec2} states the problem definition and describes the proposed framework in detail. Section \ref{sec3} presents experiments conducted on the HCP dataset. In Section \ref{sec4}, the experiments' results and discussion are given. Finally, Section \ref{sec5} concludes the paper.

\section{Related Work}\label{sec:related_work}


\subsection{Multi-Parameter Regularization}
To address the ill-posedness of the microstructure estimation problem, there are different ways to incorporate domain knowledge as regularization terms in the diffusion MRI area. These include spatial regularization methods such as sparsity \cite{cheng2015joint, schwab2018joint, yap2016multi}, low-rank \cite{lam2016fast, zhang2020acceleration, ramos2021snr} and total variation \cite{shi2016super, baust2016combined}, as well as q-space regularization \cite{hess2006q, descoteaux2007regularized} and task-specific regularization methods \cite{jonasson2007representing, savadjiev20063d, chen2023deep}. 
Specifically, \cite{chen2023deep} proposed two novel loss functions to implicitly leverage tissue microstructural characteristics by computing the second-order central moment of the DWI signal for regularization. 

Recent literature has also explored various strategies to leverage redundancy across multi-parameter MRI data. For instance, MDI \cite{ye2021multi} leverages the dimensional orthogonality of MR signals to suppress unwanted effects and streamline subsequent image processing. HD-PROST \cite{bustin2019high} introduced a patch-based reconstruction framework for accelerated 2D T1-T2 mapping and 3D magnetization transfer-weighted imaging by exploiting local and non-local redundancies across multiple contrasts. Zhao et. al \cite{zhao2024high} developed a method that integrates subspace and an adaptive generative image prior for MRI reconstruction. While these methods highlight the potential of incorporating multi-parameter priors, they are generally confined to 2D/3D contexts. In contrast, diffusion MRI inherently involves 4D signals. To address this gap, we propose a tensor-SVD-based regularization framework specifically tailored to the characteristics of high-dimensional diffusion parameter estimation. Compared with previous methods, our approach jointly captures spatial locality, inter-slice coherence, and inter-parameter correlation in a unified framework,  thereby enabling more comprehensive and effective exploitation of the multi-parameter structure in dMRI.

\begin{figure*}[!t]
\centerline{\includegraphics[width=\textwidth]{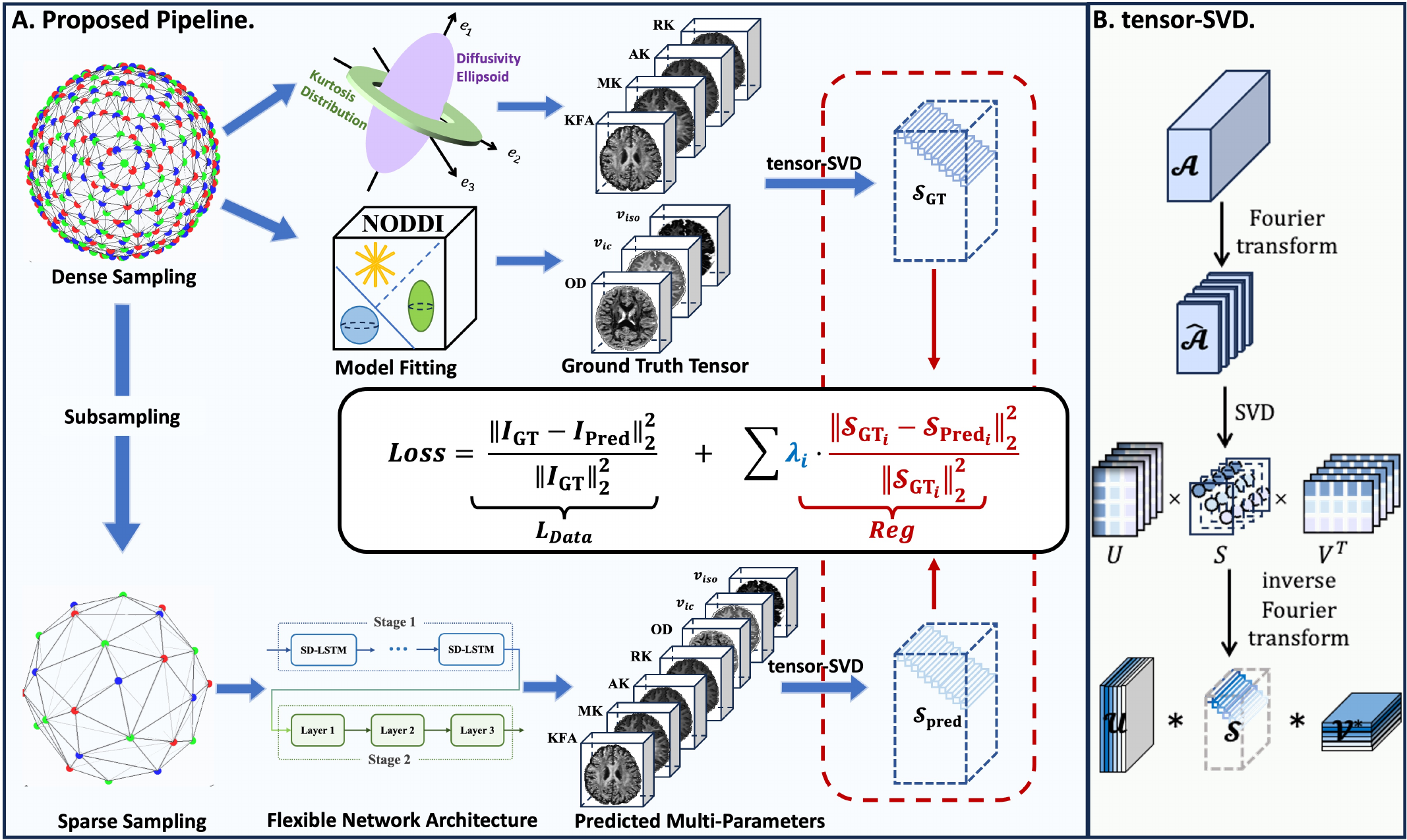}}
\caption{An illustration of the proposed DeepMpMRI framework. Left: The whole architecture consists of two branches, with the upper branch being the reference acquisition from the dense sampling and the lower branch being the network prediction from the sparse sampling. The input of the framework is sparse measurements uniformly sampled at each shell from the dense measurements, and tensor Singular Value Decomposition (t-SVD) is applied to both network prediction and reference to further align them on the tensor singular values to capture the correlation shared among multiple microstructural parameters. Right: The flowchart of tensor-SVD implementation. For detailed information, please refer to \cite{kolda2001orthogonal, kolda2009tensor, chen2009tensor, kilmer2011factorization,  lu2019tensor}.}
\label{pipeline}
\end{figure*}

\subsection{Adaptive Tuning Algorithm}
The process of hyperparameter selection is often based on trial-and-error or exhaustive strategies such as grid and random search \cite{bergstra2012random}, or on generalized cross-validation \cite{golub1979generalized}, which can be time-consuming and computationally intensive. To improve efficiency, several deep learning–based methods have been proposed for adaptive regularization parameter tuning. For example, in multi-task learning, relationships among tasks are commonly exploited to guide the weighting of task-specific loss terms \cite{kendall2018multi, guo2018dynamic}. However, since our framework does not involve multiple tasks but rather a unified task with multi-parameter estimation, such strategies are not directly applicable. 
Bilevel optimization methods (BLOs) \cite{liu2021investigating} have also been used for adaptive hyperparameter tuning, but their application is often limited due to nonconvexity and non-differentiability, making BLOs highly complicated and computationally challenging to solve \cite{afkham2021learning}. Moreover, several studies have explored adaptive learning rate algorithms via gradient-based techniques \cite{baydin2017online, chandra2022gradient, xie2024adan}. Despite promising results, they typically require computing gradients concerning all network parameters and often rely on the automatic differentiation mechanism \cite{chandra2022gradient}, which complicates their integration and increases computational overhead.
To this end, we propose a lightweight adaptive tuning algorithm based on Nesterov's accelerated gradient method, specifically designed for scenarios involving a small number or even a single regularization parameter. Our method only requires storing the previous gradient during training through manual computation of the gradient, thus being efficient in terms of both memory and time complexity. It can be implemented with just a few additional lines of code and can be easily integrated into existing frameworks.



\section{MATERIAL AND METHOD}\label{sec2}
This section provides a concise overview of the basic notations used in this paper, followed by a statement of the problem and a detailed presentation of the proposed DeepMpMRI. The proposed method is depicted in Fig. \ref{pipeline} and encompasses key elements, specifically the tensor-decomposition-based regularization (TDR) and the Nesterov-based adaptive learning algorithm (NALA).

As illustrated in Fig. \ref{pipeline}, the overall architecture consists of two branches, with the upper branch representing the reference acquisition from dense sampling, while the lower branch symbolizes the network prediction from the sparse sampling. The network input is sparse measurements uniformly sampled from the dense measurements using DMRITool \cite{cheng2015novel, cheng2017single}, and tensor Singular Value Decomposition (t-SVD) \cite{kolda2001orthogonal, kolda2009tensor, chen2009tensor} is applied to both network prediction and reference to further align them on the tensor singular values to capture the correlation shared among multiple microstructural parameters.

\subsection{Notation and Material}
This paper uses the commonly used notation where a tensor with a capital calligraphic letter $\mathbf{\mathcal{A}}$ is a multidimensional array. If $\mathbf{\mathcal{A}}$ is a tensor of order-$p$, then can be written as 
$\mathbf{\mathcal{A}} = \left( \mathbb{\mathit{a}}_{\mathbb{\mathit{i}}_{1} \mathbb{\mathit{i}}_{2} \ldots \mathbb{\mathit{i}}_\mathbb{\mathit{p}}} \right) \in \mathbb{R}^{n_{1} \times n_{2} \times \cdots \times n_{p}}$.
A matrix (order-$2$ tensor) is denoted with a bold capital letter, $\mathbb{\mathit{\textbf{A}}} \in \mathbb{R}^{n_1 \times n_2}$, a vector (order-$1$ tensor) is denoted with a bold lowercase letter, $\mathbb{\mathit{\textbf{a}}}\in\mathbb{R}^{n_1\times 1}$, and scalar (order-$0$ tensor) is denoted by a lowercase letter, $a\in R$.

There exist several well-known methods to decompose a high-order tensor $\mathbf{\mathcal{A}}$: CANDECOMP/PARAFAC (CP) \cite{hitchcock1927expression, harshman1970foundations},  Tucker \cite{tucker1963implications},  higher-order SVD (HOSVD) \cite{de2000multilinear}, Tensor-Train \cite{oseledets2011tensor}, and tensor SVD \cite{kolda2001orthogonal, kolda2009tensor, chen2009tensor}. Among these definitions, only tensor-SVD satisfies an Eckart-Young-like Theorem \cite{kilmer2011factorization, kilmer2021tensor}, which parallels that of the matrix SVD. 
{Suppose $\mathbf{\mathcal{A}}\in \mathbb{R}^{n_{1} \times n_{2} \times n_{3}}$, then it can be factorized as $\mathbf{\mathcal{A}} =  \mathbf{\mathcal{U}} * \mathbf{\mathcal{S}} * \mathbf{\mathcal{V^*}} $, where {$\mathbf{\mathcal{U}}\in \mathbb{R}^{n_{1} \times n_{1} \times n_{3}}, \mathbf{\mathcal{V}}\in \mathbb{R}^{n_{2} \times n_{2} \times n_{3}}$} are tensor orthogonal \cite{kilmer2011factorization}, {$\mathbf{\mathcal{S}}\in \mathbb{R}^{n_{1} \times n_{2} \times n_{3}}$} is f-diagonal \cite{kilmer2011factorization}.}
The specific definition and computation of the tensor-SVD are beyond the scope of this paper. For detailed information, please refer to \cite{kolda2001orthogonal, kolda2009tensor, chen2009tensor, kilmer2011factorization, lu2019tensor}. 

\subsection{Problem Definition of Microstructure parameter estimation}
By applying a set of diffusion gradients, we can acquire a vector of diffusion signals at each voxel. Each diffusion signal can be considered as a set of $ W\times H\times S$ size volumes captured in the q-space. Thus, the dMRI data are 4D signals of size $\mathbb{R}^{W\times H\times S\times D}$, where $W$, $H$, $S$, $D$ refer to the width, height, slice, and gradient directions, respectively.

We aim to estimate the multiple microstructural parameters derived from various diffusion models using sparsely sampled q-space data. Given the diffusion MRI data $\mathcal{X}\in\mathbb{R}^{W\times H\times S\times D_{Full}}$, containing the full measurements in the q-space, {we can obtain the corresponding ground-truth scalar maps $\mathcal{Y}_{GT}\in\mathbb{R}^{W\times H\times S\times N}$ by model fitting, where $N$ represents the number of desired parameters}. The network $\mathcal{F}_\theta$ parameterized by $\theta$ is designed to learn a mapping from the given sparse sampling data $\widetilde{\mathcal{X}}{\in\mathbb{R}}^{W\times H\times S\times D_{Sparse}}$ to predicted multi-parameters $\mathcal{Y}$ s.t. $\mathcal{Y} = \mathcal{F}_\theta(\widetilde{\mathcal{X}})\rightarrow \mathcal{Y}_{GT}$.

{It has been recently shown that parameter estimation for advanced models is challenging under normal experimental conditions \cite{jelescu2016degeneracy, novikov2018modeling, jelescu2020challenges}, like the NODDI \cite{zhang2012noddi} model.} 
To tackle this problem, prior knowledge was introduced as a regularization term to constrain the solution:
\begin{equation}
\min_{\theta} L_{Data}(\mathcal{F}_\theta(\widetilde{\mathcal{X}}),\mathcal{Y}_{GT})+\lambda \cdot R_{\theta}(\widetilde{\mathcal{X}})
\label{loss0}
\end{equation}
where $L_{Data}(\cdot)$ denotes the data fidelity between the network output $\mathcal{F}_\theta(\widetilde{\mathcal{X}})$ and reference $\mathcal{Y}_{GT}$, $R_{\theta}(\cdot)$ is the regularization applied to the input $\widetilde{\mathcal{X}}$. 
The hyperparameter $\lambda$ is a scalar determining how strongly the prior knowledge will be weighted during estimation.

\subsection{DeepMpMRI}
\subsubsection{Tensor-decomposition-based regularization}
To facilitate accurate and fine-grained reconstruction of microstructure, we explicitly consider the correlation of derived parameters in the overall optimization problem:
\begin{equation}
\min_{\theta}L_{Data}(\mathcal{F}_\theta(\widetilde{\mathcal{X}}),\mathcal{Y}_{GT})+\lambda \cdot R(\mathcal{F}_\theta(\widetilde{\mathcal{X}}),\mathcal{Y}_{GT})
\label{loss}
\end{equation}
here $\textit{R}(\cdot)$ is the regularization over the given predicted metrics $\mathcal{F}_\theta(\widetilde{\mathcal{X}})$. To simultaneously estimate multiple microstructural parameters from various diffusion model, a multi-head deep network is employed to explore inter-model complementary information and fuse the two diffusion representations and their derived high-level features within the network. The objective function is partly defined based on the network output to enforce the consistency between the predicted and the reference scalar maps $\mathcal{Y}_{GT}$ obtained from the full sampling. The loss term for data is defined as:

\begin{equation}
L_{Data} = \frac{\left\|\mathcal{Y}_{GT}-\mathcal{Y}\right\|_2^2}{\left\|\mathcal{Y}_{GT}\right\|_2^2}
\label{data_loss}
\end{equation}

As an extension of the Singular Value Decomposition (SVD), tensor SVD has been shown to be superior in capturing the spatial-shifting correlation beyond the typical 2D spatial domain \cite{song2020robust}. Specifically, the tensor SVD based on the tensor-product framework begins by applying a fast Fourier transform (FFT) along the third and fourth dimensions of a four-dimensional parameter tensor ($W \times H \times S \times N$) \cite{kilmer2011factorization, martin2013order}. This transformation decouples intra-slice spatial coherence and inter-parameter dependencies by converting the data from the complex spatial signal into several separate frequencies. Subsequently, matrix SVD is performed across the first two dimensions ($W \times H$) to capture local spatial correlations. Through this process, the method systematically extracts both spatial and structural coherence, as well as cross-parameter correlations, embedded in the high-dimensional tensor structure. To incorporate this correlation-aware constraint into network training, we apply the tensor SVD regularization to both the predicted multi-parameter tensor and the reference tensor. The resulting singular value tensor $\mathbf{S}$ retains the energy distribution of the data across parameters and slices, serving as a compact representation of informative variations. By jointly modeling spatial consistency and inter-parameter correlations in a unified tensor framework, the proposed regularization enhances the robustness and accuracy of multi-parametric diffusion estimation.

Referring to the definition of the tensor nuclear norm by Lu et al.\cite{lu2019tensor}, the regularization is defined as:
\begin{equation}
\begin{aligned}
    R = \frac{\left\|\text{tSVD}(\mathcal{Y}_{GT})-\text{tSVD}(\mathcal{Y})\right\|_2^2}{\left\|\text{tSVD}(\mathcal{Y}_{GT})\right\|_2^2} = \frac{\left\|\mathcal{S}_{G T}-\mathcal{S}\right\|_2^2}{\left\|\mathcal{S}_{G T}\right\|_2^2}
\end{aligned}
\label{tSVD_loss}
\end{equation}
where $\mathcal{S}$ and $\mathcal{S}_{GT}$ refer to the singular value tensors of reference $\mathcal{Y}_{GT}$ and prediction $\mathcal{Y}$ respectively.

The Eckart–Young theorem \cite{eckart1936approximation, carroll1970analysis} suggests that the principal tensor singular values encapsulate the primary information of the microstructure parameters, whereas the components associated with smaller tensor singular values may contain noise. As a result, our focus is primarily on aligning the primary tensor singular values to ensure consistency in the extraction of significant information, effectively preserving fine details while reducing noise.

The total loss is the weighted combination of the above two loss terms: 
\begin{equation}
\begin{aligned}
    L &= L_{Data} +\lambda \cdot R(\cdot) \\
    &= \frac{\left\|\mathcal{Y}_{GT}-\mathcal{Y}\right\|_2^2}{\left\|\mathcal{Y}_{GT}\right\|_2^2} + \lambda \cdot \frac{\left\|\mathcal{S}_{G T}-\mathcal{S}\right\|_2^2}{\left\|\mathcal{S}_{G T}\right\|_2^2}
\label{total_loss}
\end{aligned}
\end{equation}
where the hyperparameter $\lambda$ balances the contribution between data fitting and prior knowledge. A $\lambda$ value of 0 indicates that the prior knowledge does not influence the estimation. In this scenario, the network solely learns the mapping between diffusion-weighted images and multiple parameters from training data without considering their correlations. Conversely, as $\lambda$ increases, the optimization places greater emphasis on extracting dominant information, potentially overlooking the subtleties embedded in the raw data. 

\subsubsection{Nesterov-based Adaptive Learning Algorithm}
{The selection of the weighting hyperparameter is crucial to achieving a reasonable solution in a regularized problem.} 
Building upon the foundation laid by previous studies, we propose a \textbf{N}esterov-based hyperparameter \textbf{A}daptive \textbf{L}earning \textbf{A}lgorithm (NALA). 
Our method optimizes the network parameter $\theta$ and hyperparameter $\lambda$ alternately on the training and validation sets, respectively. Let $\lambda_t$ and $\theta_t$ be the values of $\lambda$ and $\theta$ at the step $t$. More specifically, the iterations go as follows:

\begin{equation}
\left\{
\begin{aligned}
    L(\theta,\lambda) &=L_{Data}(\theta)+\lambda \cdot R(\theta) \\
    \text{On Training Set: } \theta_{t+1} &= \arg \min_\theta{L(\theta,\lambda_t)} \\
    \text{On Validation Set: } \lambda_{t+1}&= \arg\ \min_\lambda{L(\theta_{t+1},\lambda)}
\label{iterative_update}
\end{aligned}
\right.
\end{equation}

In analogy to updating network parameter $ \theta$, $\lambda_t$ should be updated in the direction of the gradient of the loss $L(\theta_t,\lambda_t)$ concerning $\lambda_t$, scaled by another hyper-hyperparameter $\beta$. One way to compute $\frac{\partial L(\theta_{t+1},\lambda_t)}{\partial\lambda_t}$ is the direct manual computation of the partial derivative:
\begin{equation}
\frac{\partial L\left(\theta_{t+1},\lambda_t\right)}{\partial\lambda_t}=\frac{\partial\left[L_{Data}\left(\theta_{t+1}\right)+\lambda_t\cdot R(\theta_{t+1})\right]}{\partial\lambda_t}= R(\theta_{t+1})
\label{gradient}
\end{equation}
where Eq. \ref{gradient} can be obtained by observing that $L_{Data}(\theta_t)$ and $R(\theta_t)$ do not depend on $\lambda_t$. This expression lends itself to a simple and efficient implementation: simply remember the past regularization value. Similar to the Nesterov Accelerated Gradient (NAG) method, we introduce the momentum term $m$ here:
\begin{equation}
m_{t+1}=\beta \cdot m_t+R(\theta_{t+1})+\beta \cdot \left[R(\theta_{t+1})-R(\theta_t)\right]
\label{momentum}
\end{equation}
where $R(\theta_{t+1})-R(\theta_{t})$ is the differential of the gradient, which approximates the second-order derivative of the objective function. Thus, the momentum term now averages the past search directions $m_t$, the current stochastic gradient $R_t$, and the approximate second-order derivative to determine the search direction $m_{t+1}$. {Previous studies \cite{Nesterov1983AMF, nesterov2013introductory} have demonstrated that the NAG can theoretically achieve faster convergence rates by using gradients at an extrapolation point of the current solution. This foresight enables better utilization of the curvature information and greater robustness \cite{xie2024adan}. By leveraging insights from it, we introduce the update rule here:}
\begin{equation}
\begin{aligned}
\lambda_{t+1}&=\lambda_t-\alpha\cdot m_{t+1}\\
&=\lambda_t-\alpha\beta\cdot m_t-\alpha\cdot R(\theta_{t+1})-\alpha\beta\cdot\left[R(\theta_{t+1})-R(\theta_{t})\right]
\label{momentum_lambda}
\end{aligned}
\end{equation}

During the initial training phase, a significant disparity exists between the predicted and reference tensors. As a result, $\lambda$, determined by Eq. \ref{momentum_lambda}, assumes a relatively large value, which facilitates the acceleration of network training. As the training process nears completion, the disparity between the predicted and reference becomes smaller, leading to a gradual decrease in lambda towards convergence. 
\cite{pereverzev2005adaptive} studied the possibility of using the structure of regularization error to adapt the regularization parameter and offered a theoretical analysis in nonlinear ill-posed problems. Our proposed algorithm operates with minimal memory usage and minimal computational overhead, requiring only a single extra past regularization copy and only one step to update.


\subsection{Backbone Network}
Note that DeepMpMRI is a flexible framework that can utilize any network architecture suitable for the task. 
Here, the Hybrid Graph Transformer (HGT) \cite{chen2022hybrid} is employed as our backbone. 
HGT utilizes the joint x-q space information with a hybrid architecture consisting of a GNN and a transformer from undersampled DMRI data for accurate tissue microstructure estimation. 
In Section \ref{backbone_section}, we investigated the effectiveness when DeepMpMRI incorporates different backbone networks.





\begin{table}[ht]
\centering
\caption{The quantitative results were obtained on the HCP dataset with 6 diffusion directions per shell at b-values of 1000, 2000, and 3000 $s/{mm}^2$ compared to Model Fitting (MF), q-DL, U-Net, MESC-SD, HGT, and Ours in terms of PSNR, SSIM, and nRMSE. The best results are in \textbf{bold}.} 
\label{tab1}
\renewcommand{\arraystretch}{1} 
\resizebox{\textwidth}{!}{
\tiny 
\begin{tabular}{c|cccccccc|cccccccc}
\hline
\multicolumn{1}{c|}{Methods} &\multicolumn{8}{c|}{PSNR$\uparrow$} &\multicolumn{8}{c}{SSIM$\uparrow$}
\\ \cline{2-17}
\multicolumn{1}{c|}{(18 DWIs)} & KFA & MK & AK & RK & OD & $V_{ic}$ & $V_{iso}$ & ALL & KFA & MK & AK & RK & OD & $V_{ic}$ & $V_{iso}$ & ALL\\
\hline
MF & 14.1646 & 7.7494 & 8.5622 & 6.4807 & 18.7080 & 18.8891 & 25.9996 & $ 10.6679 \pm 0.47$ & 0.7107 & 0.7036 & 0.7037 & 0.7036 & 0.8510 & 0.8771 & 0.9370 & $0.7838 \pm 0.0241$ \\
q-DL & 25.9322 & 20.9010 & 21.0300 & 19.0106 & 25.7907 & 28.5115 & 31.5925 & $22.8977 \pm 0.47$ & 0.9412 & 0.9602 & 0.9323 & 0.9562 & 0.9495 & 0.9762 & 0.9701 & $0.9546 \pm 0.0055$ \\
U-Net & 26.1232 & 21.4837 & 21.3266 & 19.6591 & 25.8229 & 27.3658 & 31.9325 & $23.2782 \pm 0.43$ & 0.9424 & 0.9710 & 0.9359 & 0.9639 & 0.9455 & 0.9682 & 0.9722 & $0.9570 \pm 0.0048$ \\
MESC-SD & 26.4769 & 21.4655 & 21.3842 & 19.7558 & 26.9755 & 28.9045 & 32.9509 & $23.5240 \pm 0.46$ & 0.9499 & 0.9712 & 0.9419 & 0.9662 & 0.9602 & 0.9761 & 0.9761 & $0.9720 \pm 0.0043$ \\
HGT & 27.1980 & 22.2145 & 22.0147 & 20.4989 & 27.8408 & 30.5431 & 34.2478 & $24.2910 \pm 0.46$ & 0.9563 & 0.9767 & 0.9480 & 0.9711 & 0.9674 & 0.9825 & 0.9813 & $0.9691 \pm 0.0035$ \\
\textbf{Ours} & \textbf{27.3180} & \textbf{22.3864} & \textbf{22.1501} & \textbf{20.6378} & \textbf{27.9499} & \textbf{30.9029} & \textbf{34.4643} & $\textbf{24.4422} \pm \textbf{0.45}$ & \textbf{0.9579} & \textbf{0.9779} & \textbf{0.9496} & \textbf{0.9726} & \textbf{0.9681} & \textbf{0.9834} & \textbf{0.9821} & $\textbf{0.9702} \pm \textbf{0.0034}$ \\
\hline
\end{tabular}}
\end{table}

\begin{figure*}[!t]
\centerline{\includegraphics[width=0.9\textwidth]{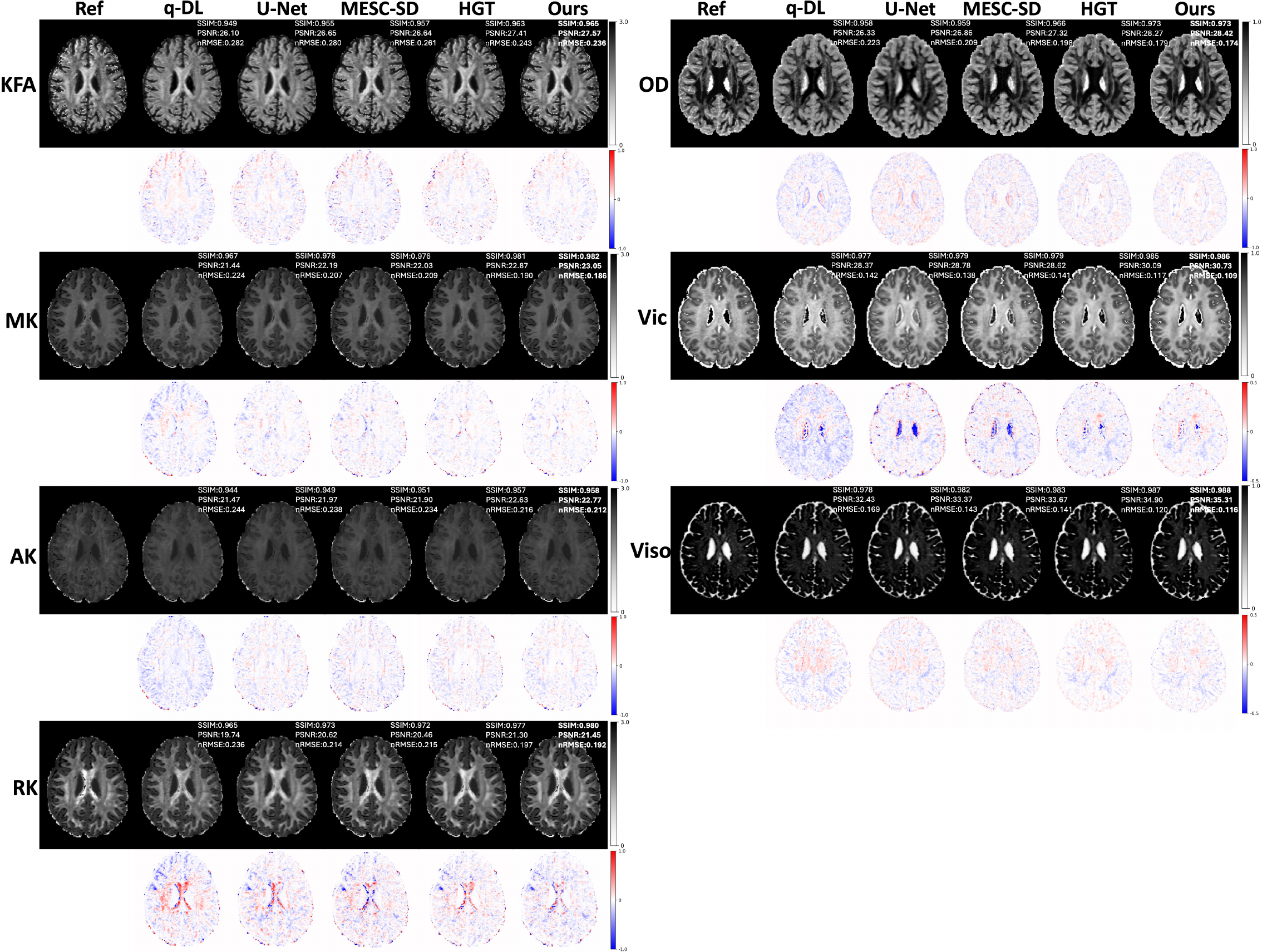}}
\caption{The reference, estimated parameters {(KFA, MK, AK, RK, OD, $V_{ic}$, $V_{iso}$)} and corresponding error maps based on Model Fitting (MF), q\_DL, U-Net, MESC-SD, and Ours in a test subject with 6 diffusion directions per shell at b-values of 1000, 2000, and 3000 $s/{mm}^2$.}
\label{main_visual}
\end{figure*}

\section{Experiments}\label{sec3}
In this section, we introduce the dataset and the compared methods used in our experiments, followed by the implementation details and experimental settings. 
\subsection{Datasets}
\subsubsection{The HCP Dataset}
Pre-processed whole-brain diffusion MRI data from the publicly available Human Connectome Project (HCP) Young Adult dataset were used for this study \cite{van2013wu}. We randomly chose 111 subjects, 60 of whom were used for training, 17 for validation, and the remaining 34 scans for testing. Diffusion MRI data were acquired at 1.25 mm isotropic resolution with four b-values ($0, 1000, 2000, 3000 s/{mm}^2$). For each non-zero b-value, 90 DWI volumes along uniformly distributed diffusion-encoding directions were acquired. 
\subsubsection{The Alzheimer's Disease Dataset}
{The dMRI scans in the Alzheimer's disease (AD) Dataset were acquired on a GE Premier scanner with an isotropic spatial resolution of 1.7 mm and 270 diffusion gradients (90 diffusion gradients on each of the three shells $b=1000, 2000, 3000 s/{mm}^2$) \cite{li2024deep, qin2021multimodal, liu2022volumetric}. The data acquisition was approved by the institutional review board of Beijing Tiantan Hospital, Capital Medical University. Preprocessing has been performed for distortion and motion correction. This dataset used for analyzing brain tissue changes associated with AD comprised 8 patients and 9 healthy control (HC) subjects.}

\subsection{Experimental Settings}
To obtain the training data, DWI volumes acquired along 6 uniform diffusion-encoding directions on each of the shells $b=1000,\ 2000,\ 3000s/{mm}^2$ were selected for each dMRI scan using DMRITool \cite{cheng2017single}. 
To obtain the reference DKI metrics, model fitting was performed on all the diffusion data using weighted linear squares implemented in the DIPY\footnote{https://github.com/dipy} software package to derive the diffusion kurtosis, including KFA, MK, AK, and RK. The three scalar tissue microstructure measures in the NODDI model were considered, which are the intra-cellular volume fraction $V_{ic}$, cerebrospinal fluid (CSF) volume fraction $V_{iso}$, and orientation dispersion (OD). The gold standard NODDI-derived metrics were computed using the AMICO algorithm\cite{daducci2015accelerated} with the dense sampling of 270 diffusion gradients. 

The neural network was implemented using the PyTorch library (codes will be available online upon acceptance of the paper). The performance is evaluated quantitatively by calculating the commonly used metrics of peak signal-to-noise ratio (PSNR), structural similarity index measure (SSIM), normalized root mean square error (nRMSE), and dice similarity coefficient.

\subsection{Compared Methods and Implementations}
Our method was evaluated both qualitatively and quantitatively, and compared to conventional model-fitting algorithms (implemented by DIPY for DKI and AMICO \cite{daducci2015accelerated} for NODDI), as well as deep learning-based approaches. The deep learning methods included the MLP proposed by Golkov et al. \cite{golkov2016q}, the U-Net introduced by \cite{gibbons2019simultaneous}, the MESC-SD model from Ye et al. \cite{ye2020improved}, and the HGT framework developed by Chen et al. \cite{chen2023deep}.
The pioneer DL-based microstructure estimation method q\_DL \cite{golkov2016q} proposed using an MLP to perform voxel-wise estimation without considering the neighborhood information. 2D U-Net-based method \cite{gibbons2019simultaneous} and HGT \cite{chen2022hybrid} reshape the raw data into slices by flattening the other two dimensions and performing slice-wise estimation. 
MESC-SD \cite{ye2020improved} predicts the tissue microstructure in the center voxel by dividing diffusion signals into image patches. Our implementations are based on the released codes.

\begin{figure}[!t]
\centerline{\includegraphics[width=0.5\textwidth]{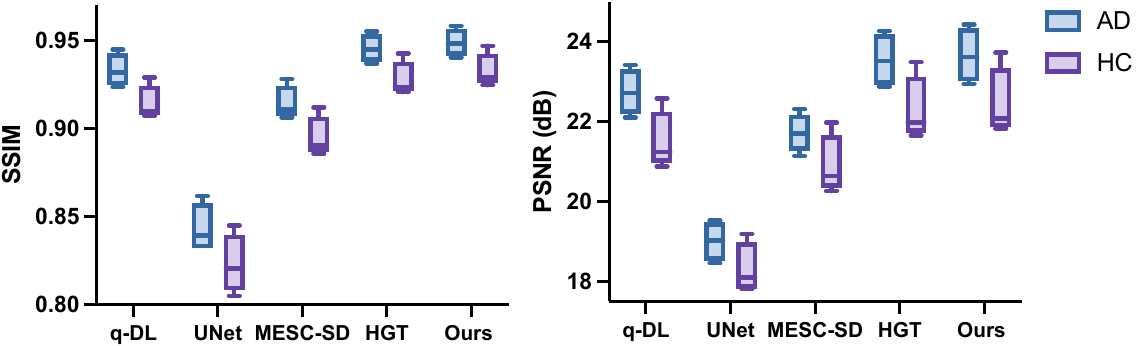}}
\caption{{Quantitative comparison of different DL-based methods on the AD dataset. The results were obtained with 6 diffusion directions per shell at b-values of $1000,\ 2000,\ 3000s/{mm}^2$.}}
\label{ad_psnr&ssim}
\end{figure}

\section{Results}\label{sec4}
\subsection{Comparison of state-of-the-art methods}
We evaluate the performance of DeepMpMRI through a comparative analysis with the backbone method and three other state-of-the-art methods. The experimental results are summarized in Table \ref{tab1}, where it can be observed that DeepMpMRI outperforms all other methods in terms of PSNR, SSIM, and nRMSE metrics. Specifically, compared to HGT with the same network architecture, DeepMpMRI significantly enhances performance across all microstructural parameters, demonstrating its effectiveness in simultaneously predicting multiple parameters from various diffusion models. 

\begin{figure}[!t]
\centerline{\includegraphics[width=0.8\textwidth]{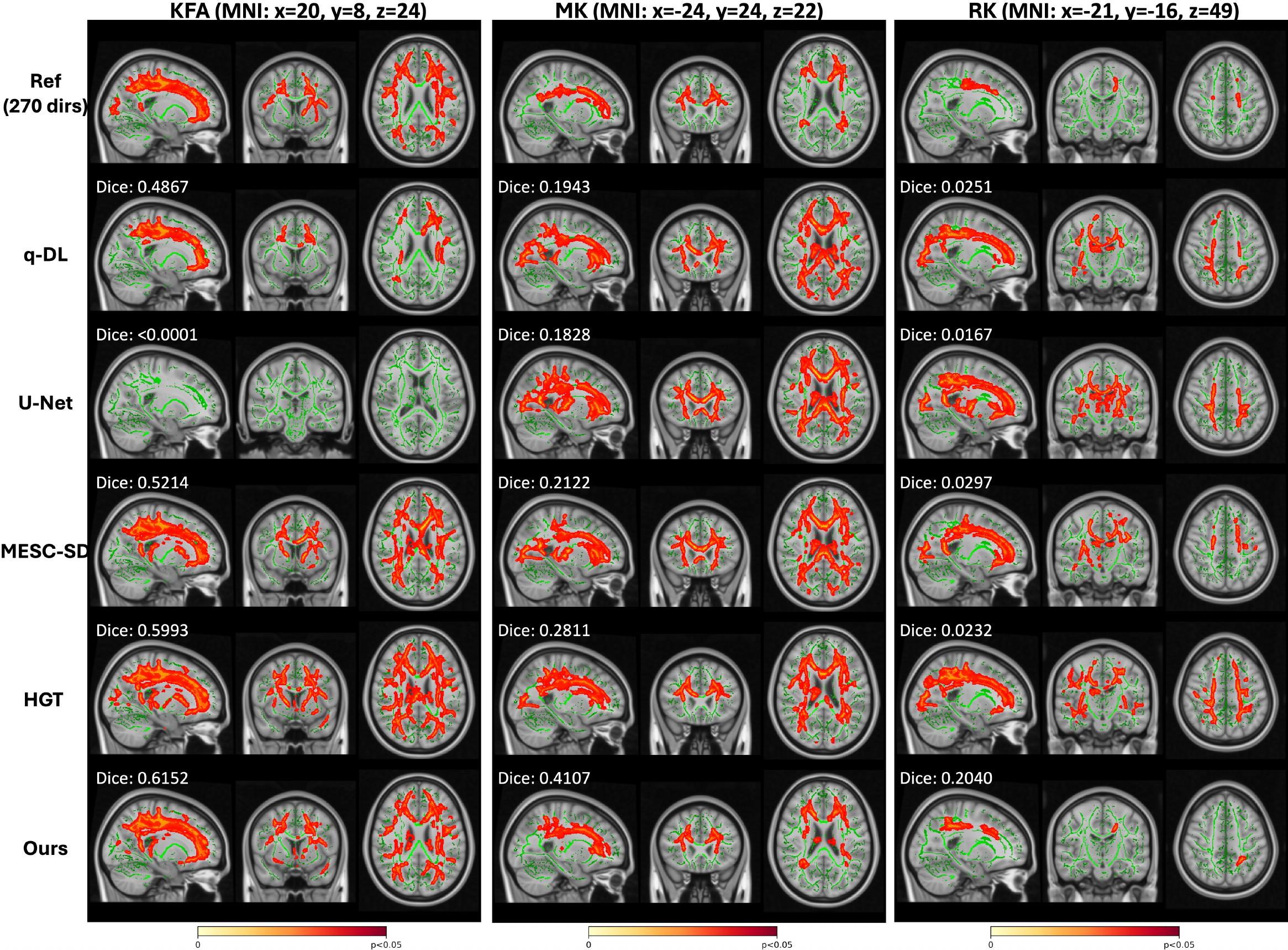}}
\caption{{Tract-based spatial statistics (TBSS) analysis revealed significant differences ($p < 0.05$) in multiple parameter values associated with changes in brain white matter for AD patients compared to healthy controls.}}
\label{AD_HC}
\end{figure}

Fig. \ref{main_visual} displays cross-sectional views of the estimated microstructure maps for a representative test subject, alongside the reference and corresponding error maps for comparison. Due to the failure of traditional model fitting methods to estimate DKI with only 18 diffusion directions, we do not include their results in the visual comparison. As shown, the simplest DL-based method, q-DL, exhibits relatively substantial estimation errors under sparse sampling conditions. The U-Net results, while smoother, suffer from excessive over-smoothing that leads to a loss of important textural details. 
We employ the HGT \cite{chen2022hybrid} as our backbone, recognized as one of the leading open-source methods for microstructure estimation. When combined with our method, it produces high-fidelity results and enhanced preservation of fine details.

\subsection{Validation on Alzheimer's Disease Dataset}
{
To validate the capability of our method to reliably identify disease-related tissue microstructure changes based on clinically feasible dMRI scans, we conducted additional experiments using a heterogeneous Alzheimer's disease dataset. Fig. \ref{ad_psnr&ssim} shows the quantitative comparisons of different DL-based methods after fine-tuning. Our method consistently achieves the lowest estimation error across both AD and HC groups. Furthermore, we employed Tract-Based Spatial Statistics (TBSS) to identify group-wise differences ($p<0.05$) in white matter changes between AD and the HC subjects across multiple parameters. Specifically, in Fig. \ref{AD_HC}, we present the regions of significant group-wise differences identified by different methods using only 18 diffusion directions. To quantitatively assess the similarity between these regions and those derived from dense sampling, we also calculated the Dice coefficient. Our results show that our method achieves superior performance both in terms of the Dice score and the visual agreement with dense sampling, highlighting the clinical relevance and generalizability of our method.}

\begin{figure}[!t]
\centerline{\includegraphics[width=0.5\textwidth]{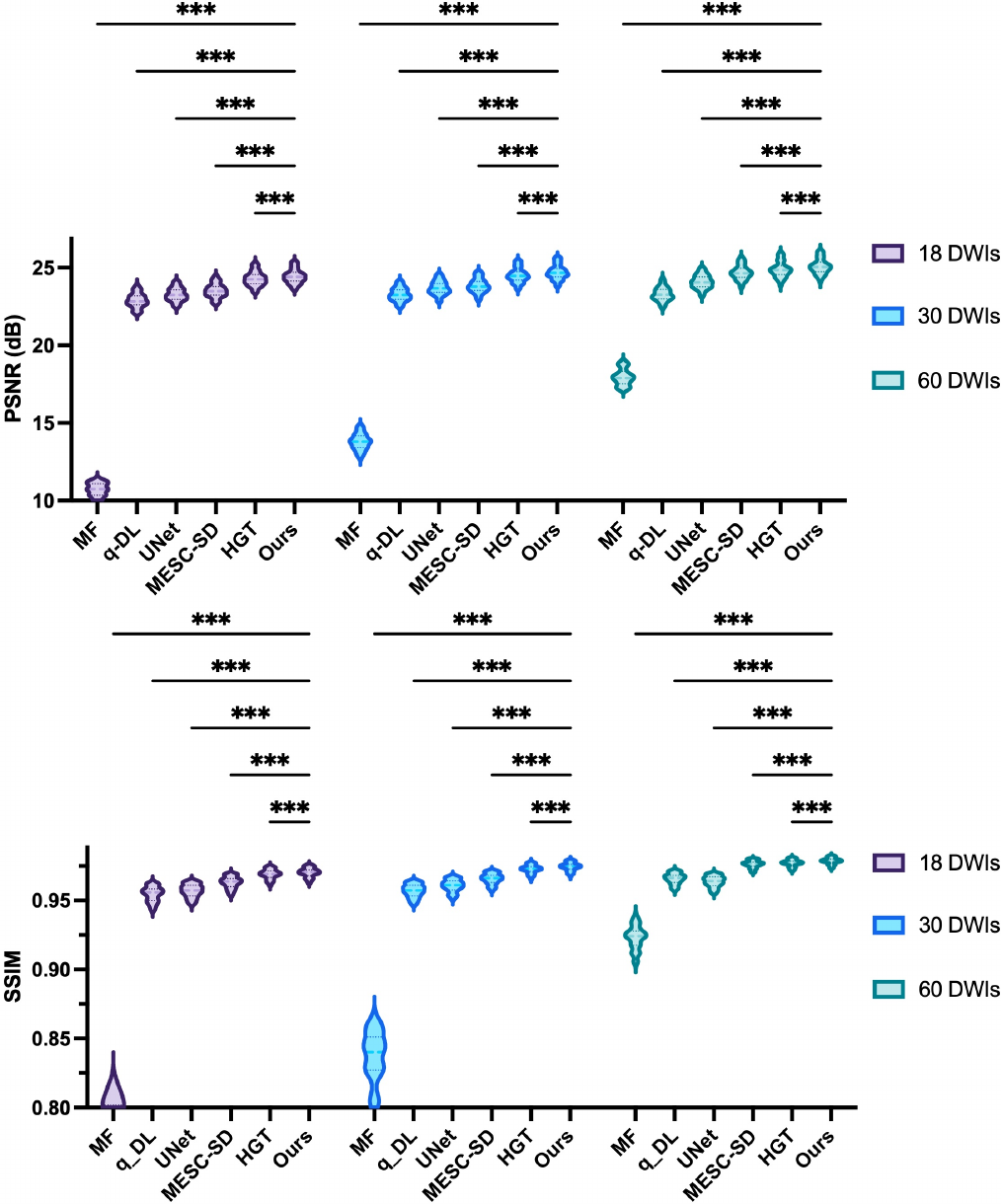}}
\caption{{Quantitative and statistical results via SSIM and PSNR on HCP Dataset. The quantitative results were obtained on four different q-space undersampling patterns compared to Model Fitting (MF), q-DL, U-Net, MESC-SD, and HGT. The $p$-value is described in the figure, where *** represents $p < 0.001$ and ** represents $p < 0.01$.}}
\label{ssim&psnr}
\end{figure}

\subsection{Comparison of different sampling rates}
To further investigate the effectiveness of our method on different sampling rates, three additional q-space undersampling patterns were considered. Consistent with the selection of 18 diffusion gradients, we utilized 30 and 60 diffusion gradients, respectively. These gradients consisted of 10 and 20 DWI volumes, acquired along uniform diffusion-decoding directions on each shell with b values of 1000, 2000, and 3000 $s/{mm}^2$. Subsequently, the proposed method and other state-of-the-art techniques were applied to all three cases. Our method achieved high-fidelity microstructure estimation with acceleration factors of 4.5-15$\times$ compared to the dense sampling of a total of 270 diffusion gradients. 
Fig. \ref{ssim&psnr} also shows the quantitative and statistical results compared between DeepMpMRI and the competing methods using paired Student's t-tests. The consistent superior performance of our method across all subsampling patterns confirmed the superiority of DeepMpMRI.

\begin{table}[ht]
\centering
\caption{{Quantitative evaluation of denoising performance using synthetic 18 DWIs on the HCP dataset with varying levels of {spatially varying non-stationary Rician noise}. The best results are in \textbf{bold}.}}\label{denoise_tab}
\resizebox{0.8\textwidth}{!}{
\begin{tabular}{c|ccc|ccc|ccc}
\hline
Methods                        & \multicolumn{3}{c|}{2.5\%}                                              & \multicolumn{3}{c|}{5\%}                                       & \multicolumn{3}{c}{7.5\%}                                               \\ \cline{2-10} 
\multicolumn{1}{l|}{(18 DWIs)} & \multicolumn{1}{c|}{PSNR} & \multicolumn{1}{c|}{SSIM} & nRMSE           & \multicolumn{1}{c|}{PSNR}             & \multicolumn{1}{c|}{SSIM} & nRMSE           & \multicolumn{1}{c|}{PSNR} & \multicolumn{1}{c|}{SSIM} & nRMSE           \\ \hline
q-DL                           & 22.5685                   & 0.9507                    & 0.2335          & 21.7719          & 0.9402                    & 0.2532          & 20.8840                   & 0.9267                    & 0.2763          \\
U-net                          & 23.2227                   & 0.9562                    & 0.2166          & 22.9856          & 0.9531                    & 0.2207          & 22.5731                   & 0.9482                    & \textbf{0.2283} \\
MESC-SD                        & 23.4022                   & 0.9615                    & 0.2106          & 22.9794          & 0.9565                    & 0.2197          & 22.0948                   & 0.9455                    & 0.2387          \\
HGT                            & 21.6346                   & 0.9569                    & 0.2666          & 20.4553          & 0.9397                    & 0.3092          & 19.6379                   & 0.9225                    & 0.3404          \\
Ours                           & 21.7897                   & 0.9590                    & 0.2607          & 20.8540          & 0.9440                    & 0.2934          & 20.1277                   & 0.9284                    & 0.3186          \\
\textbf{Ours (Truncated)}       & \textbf{23.7961}          & \textbf{0.9651}           & \textbf{0.2016} & \textbf{23.2441} & \textbf{0.9587}           & \textbf{0.2146} & \textbf{22.5822}          & \textbf{0.9499}           & 0.2310          \\ \hline
\end{tabular}}
\end{table}

\begin{figure*}[!t]
\centerline{\includegraphics[width=0.8\textwidth]{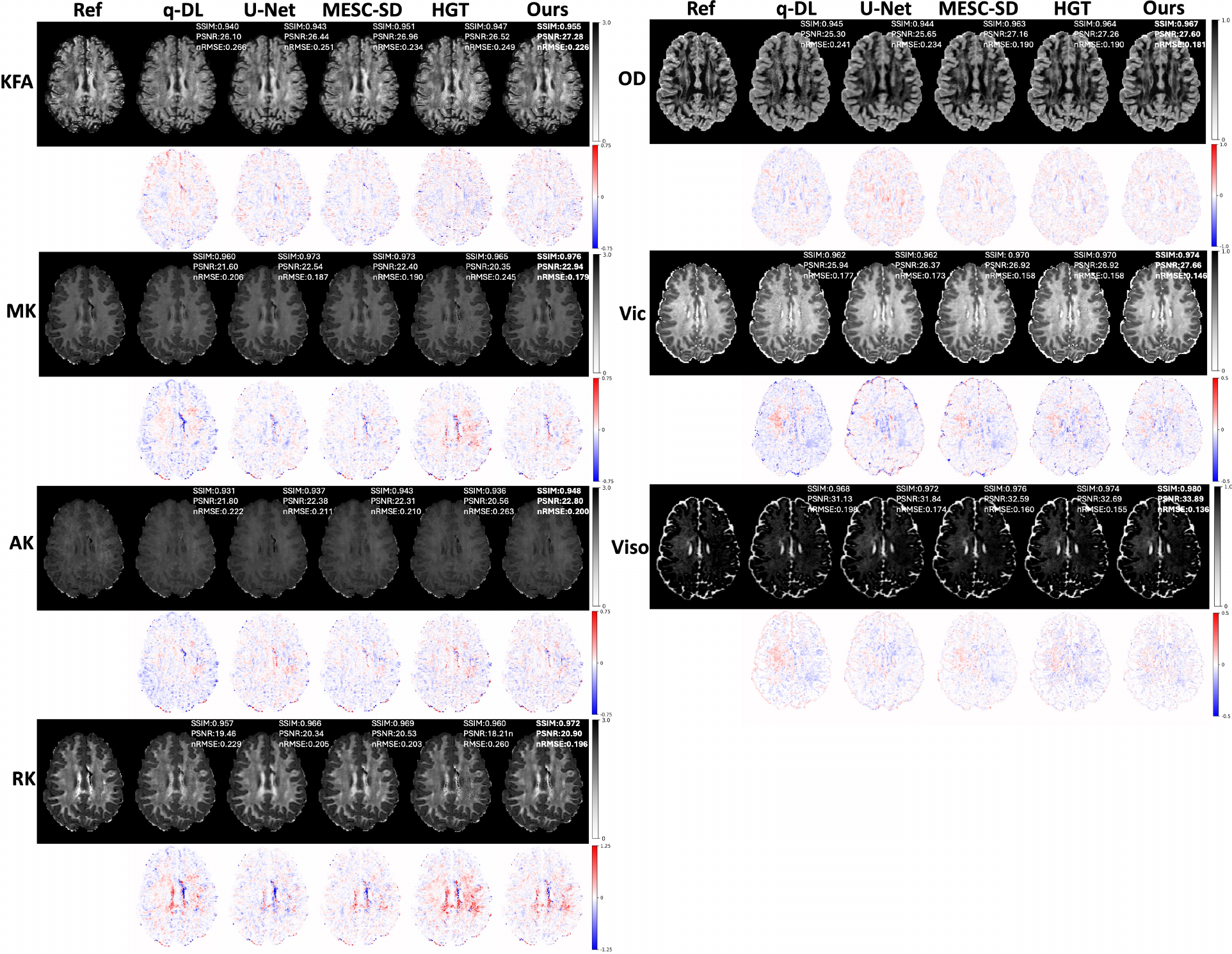}}
\caption{{The reference, estimated parameters {(KFA, MK, AK, RK, OD, $V_{ic}$, $V_{iso}$)} and corresponding error maps based on Model Fitting (MF), q\_DL, U-Net, MESC-SD, HGT, and Ours in synthetic data with $\sigma=0.025$ with 6 diffusion directions per shell at b-values of 1000, 2000, and 3000 $s/{mm}^2$.}}
\label{denoised_fig}
\end{figure*}

\subsection{Comparison of different noise tlevels}
\label{denoise_sec}
Following recent works and considering HCP noise properties \cite{aja2016statistical, chen2019noise}, we adopted the non-stationary Rician noise model to simulate spatially varying noisy data here. We also investigated the effectiveness of singular-value tensor truncation, where truncation refers explicitly to the discarding of the smallest one for each frontal slice \cite{kilmer2013third}.
Table \ref{denoise_tab} presents a comparative analysis of various methods under different noise levels. It can be observed that our method consistently outperforms other DL-based approaches. Notably, as the noise level increases, the performance gap between our method and the original HGT \cite{chen2022hybrid} method widens, especially when truncation is applied, PSNR increases by 9.99\%, 13.63\%, and 14.99\% across increasing noisy levels.
The results indicate that incorporating our method consistently improves performance under varying noise types and noise levels. With our method integrated, all evaluation metrics show a significant improvement. Specifically, PSNR increases by 9.99\%, 13.63\%, and 14.99\% over the backbone method HGT across different noise levels. Furthermore, as observed in Fig. \ref{denoised_fig}, our method retains high-fidelity microstructural parameter estimation under noisy conditions, which demonstrates that our framework is highly robust to realistic noise and maintains accurate microstructural parameter estimation even under severe degradation.

\begin{figure}[!t]
\centerline{\includegraphics[width=0.5\textwidth]{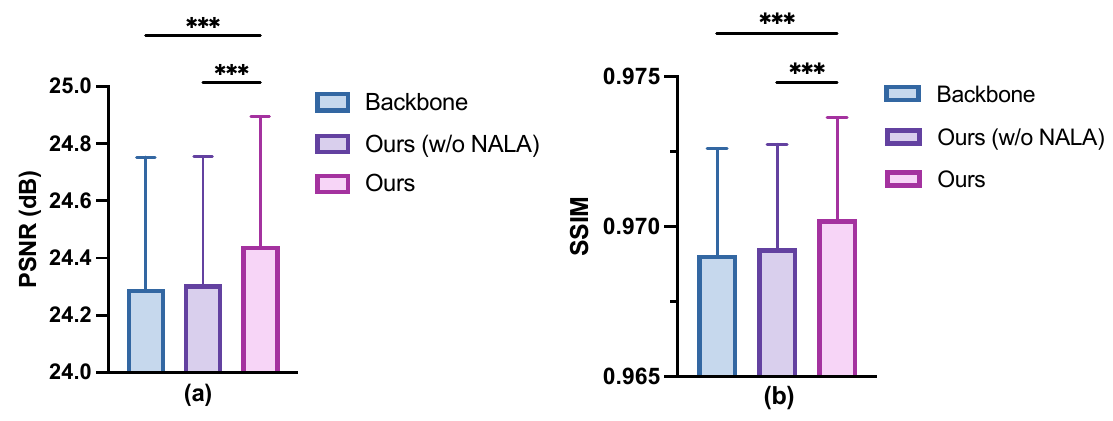}}
\caption{{Quantitative and statistical results of verification experiment of TDR and NALA. (w/o NALA) indicates DeepMpMRI without the Nesterov-based adaptive learning algorithm. The $p$-value is described in the figure, where *** represents $p < 0.001$.}}
\label{ablation_t_test}
\end{figure}

\begin{figure}[!t]
\centerline{\includegraphics[width=0.5\textwidth]{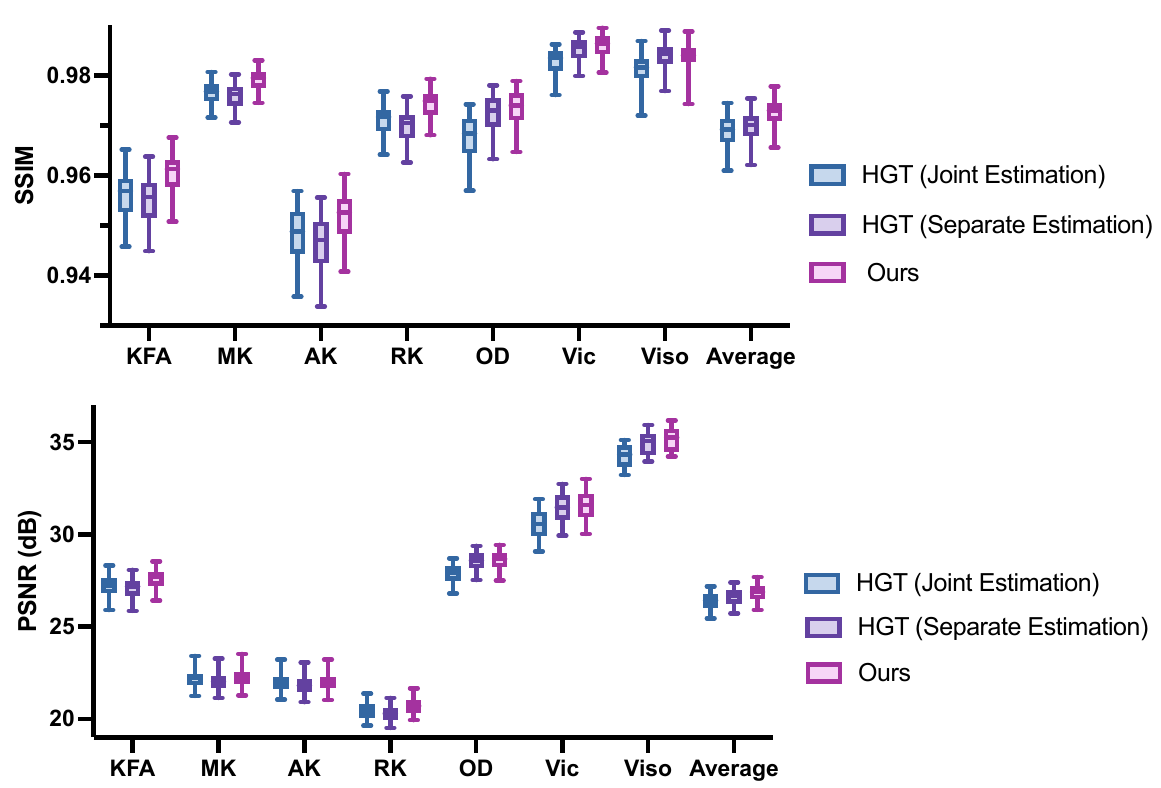}}
\caption{{Quantitative and statistical results of verification experiment of TDR and NALA. (w/o NALA) indicates DeepMpMRI without the Nesterov-based adaptive learning algorithm. The $p$-value is described in the figure, where *** represents $p < 0.001$.}}
\label{joint}
\end{figure}

\subsection{Ablation Study}
\subsubsection{Verification of TDR and NALA}
We perform an extensive ablation study to investigate the effectiveness of the tensor-decomposition-based regularization (TDR) module and Nesterov-based adaptive learning algorithm (NALA). The ablation study is completed under the condition of a total of 18 gradients (6 diffusion directions per shell at b-values of 1000, 2000, and 3000 $s/{mm}^2$). As illustrated in Fig. \ref{ablation_t_test}, the average values of PSNR and SSIM achieved by integrating both NALA and TDR are the highest among the three variants.

\subsubsection{Verification of Efficiency}
To investigate the efficiency of our method, we conduct additional experiments to compare joint estimation with separate estimation. The results in Fig. \ref{joint} indicate that, without our proposed method, separate estimation performs better, but it requires training multiple models, which significantly increases the training time and computational burden. However, incorporating our method, joint estimation with the same network architecture achieves superior performance while reducing computational costs. Our proposed method provides a potential solution for fast, accurate, and efficient microstructure estimation. 

\subsubsection{Influence of $\alpha$ and $\lambda$}
{
As suggested by Adam \cite{kingma2014adam}, we fixed the $\beta=0.9$, and since $m_0$ refers to the initial momentum/velocity, we set $m_0=0$. We presented a comparison of experimental results under fixed initial $\lambda$ with changing $\alpha$. Under the same initial conditions ($\lambda=0.1, m_0=0, \beta=0.9$), we adjusted the learning rate $\alpha$ and observed the optimization process of the $\lambda$. Fig. \ref{Lambda_Alpha} shows that the learning rate $\alpha$ affects the update step size of $\lambda$: a higher $\alpha$ means that $\lambda$ changes more significantly in each iteration, while a lower $\alpha$ results in a more gradual update. Therefore, we suggest that a higher initial $\lambda$ necessitates a relatively larger $\alpha$. For this study, the initial value of $\lambda$ was set at 0.1 and the $\alpha=5e^{-4}$.
}

\begin{figure}[!t]
\centerline{\includegraphics[width=0.5\textwidth]{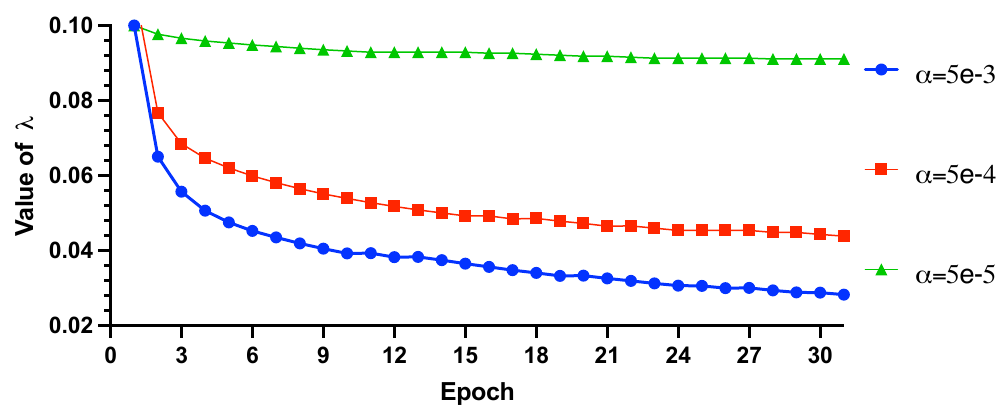}}
\caption{{Lambda learning process curve under the same initial value and different learning rates.}}
\label{Lambda_Alpha}
\end{figure}

\begin{figure}[!t]
\centerline{\includegraphics[width=0.5\textwidth]{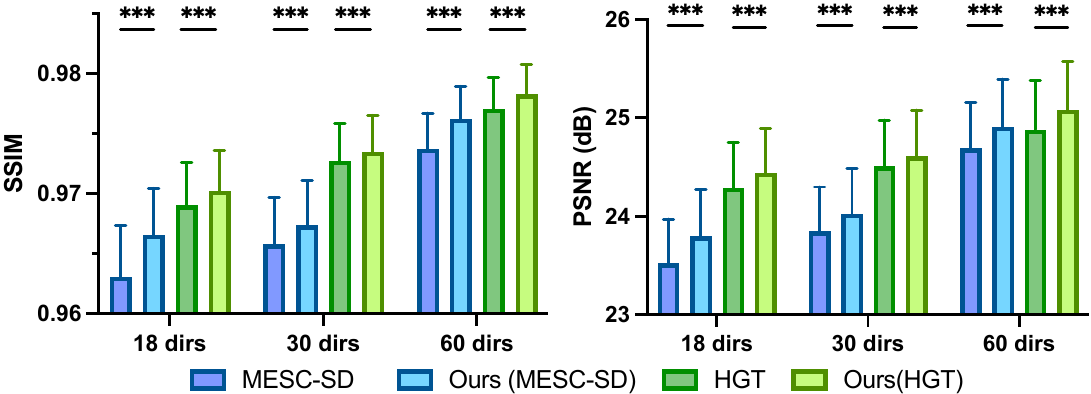}}
\caption{Average microstructure estimation and statistical results using different backbone methods (MESC-SD and HGT), in terms of SSIM and PSNR, under different q-space undersampling patterns. {The $p$-value is described in the figure, where *** represents $p < 0.001$ and ** represents $p < 0.01$.}}
\label{backbone}
\end{figure}

\section{Discussion}
\label{sec5}
\subsection{Discussion on the Different Backbone Methods}
\label{backbone_section}
As previously noted, DeepMpMRI is a highly flexible framework that can employ any network architecture as the backbone, provided the output can be reshaped to tensor form. 
Fig. \ref{backbone} presents the results obtained using various backbone networks across various undersampling patterns, with MESC-SD \cite{ye2020improved} and HGT \cite{chen2022hybrid} chosen as examples. The results demonstrate that, regardless of the backbone network utilized, the integration of the proposed DeepMpMRI consistently yields better results, further validating the effectiveness of the proposed method. 

\begin{figure}[!t]
\centerline{\includegraphics[width=0.5\textwidth]{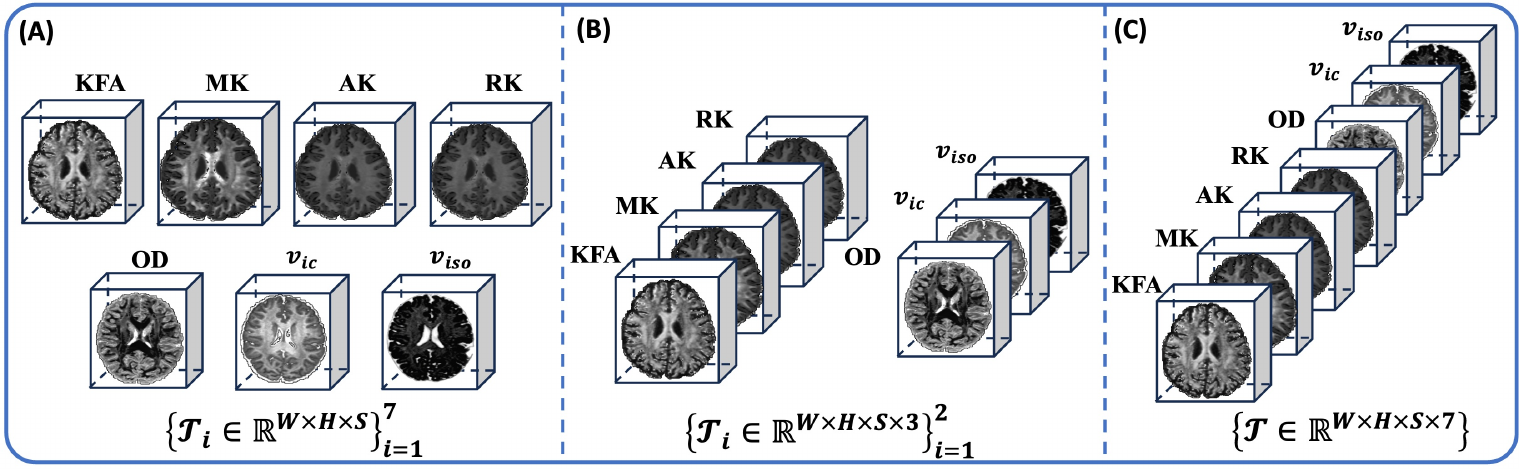}}
\caption{{The multi-parameter tensor can be decomposed in three ways: (A) decomposing the seven parameters separately; (B) combining DKI and NODDI parameters into two groups, and (C) merging all parameters as a whole tensor.}}
\label{Tensor}
\end{figure}

\subsection{Discussion on the dimensions of the multi-parameter Tensor}
The high dimensionality of multiple parameters presents various opportunities for tensor decomposition. In this section, we investigate the influence of tensor dimensions on the proposed method. We conducted a comparative study on three decomposition ways, as illustrated in Fig. \ref{Tensor}: decomposing the seven parameters into individual tensors, grouping the DKI and NODDI parameters into two sets, and merging all seven parameters into a whole tensor. The findings in Fig. \ref{tensordim} indicate that the optimal performance is achieved by merging the seven parameters, emphasizing the effectiveness of DeepMpMRI in leveraging the correlations among multiple parameters. The statistical results presented in Fig. \ref{tensordim} demonstrate that our method shows significant improvements over the original HGT, regardless of the decomposition type employed, thereby underscoring the advantages of utilizing the redundancy inherent in high-dimensional data.

\subsection{Future Work}
As demonstrated in Table \ref{denoise_tab}, the truncation operation plays a critical role in handling noisy data. In this study, we evaluated no truncation and fixed-rank truncation schemes. Adaptive truncation strategies may further optimize performance across datasets with heterogeneous noise characteristics. Future work will focus on developing adaptive rank selection algorithms to automatically determine the optimal truncation rank based on noise patterns.
Furthermore, our experiments indicate that the newly introduced hyperparameter $\alpha$ also influences the performance. To address this issue, future research will explore the recursive implementations of adaptive schemes that also dynamically adjust those hyper-hyperparameters during training. Such recursive frameworks have demonstrated their effectiveness in prior work \cite{baydin2017online, chandra2022gradient}. Incorporating such recursive techniques into our optimization pipeline may improve convergence stability and reduce sensitivity to the initial human-chosen parameters.
Beyond methodological refinements, we intend to extend our framework to provide uncertainties alongside deterministic predictions. This enhancement could inform risk-aware clinical decision-making and improve the reliability of downstream analyses, particularly in diagnostic applications where error margins critically impact patient outcomes.

\begin{figure}[!t]
\centerline{\includegraphics[width=0.5\textwidth]{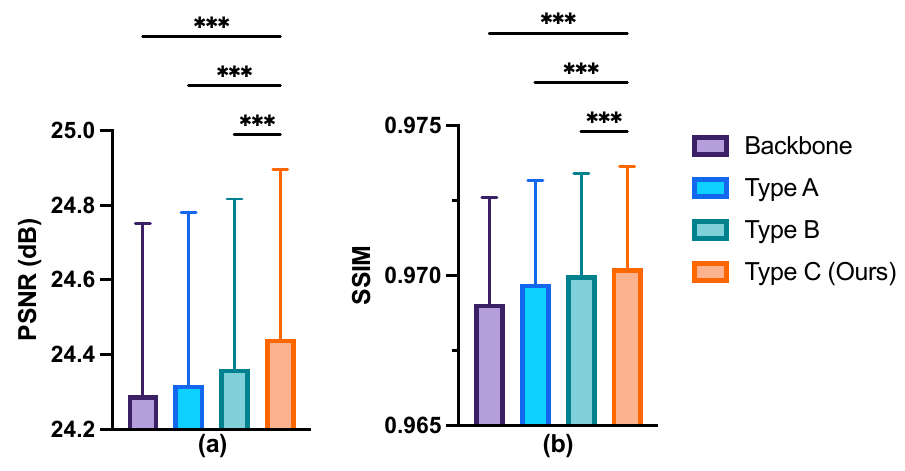}}
\caption{Quantitative comparison of three different decomposition methods based on SSIM and PSNR. The results were obtained with 6 diffusion directions per shell at b-values of 1000, 2000, and 3000 $s/{mm}^2$. The $p$-value is described in the figure, where *** represents $p < 0.001$, ** represents $p < 0.01$ and * represents $p < 0.05$.}
\label{tensordim}
\end{figure}

\section{Conclusions}\label{sec5}
In this work, we proposed DeepMpMRI, a deep learning framework for the simultaneous estimation of multiple microstructural parameters derived from various diffusion models using sparsely sampled q-space data. To improve accuracy and robustness under noisy and sparsely sampled conditions, we introduced a tensor-decomposition-based regularization that effectively leverages shared anatomical structures and diffusion characteristics across multiple parameters. Additionally, we designed a Nesterov-based adaptive learning strategy to dynamically adjust the regularization parameter, enabling efficient and reliable hyperparameter tuning during training.
{
Extensive evaluations on the HCP dataset under various undersampling patterns demonstrate that our method consistently outperforms state-of-the-art approaches in both quantitative accuracy and visual quality. Notably, even under high noise contamination, DeepMpMRI still achieves fine-grained and biologically meaningful microstructure estimation. More importantly, our method is capable of detecting disease-relevant microstructural alterations in white matter using sparse measurements, highlighting its clinical utility and robustness in realistic scenarios.
Overall, DeepMpMRI represents a novel and powerful solution for fast, efficient, and high-fidelity multiple microstructure imaging, offering a promising direction for future research and translational neuroimaging applications.}

\bmsection*{Author contributions}
Wenxin Fan, Jian Cheng and Shanshan Wang contributed to the theoretical development and experimental design. Wenxin Fan, Juan Zou, and Ruoyou Wu contributed to the analysis and interpretation of data. Wenxin Fan and Ruoyou Wu contributed to the original draft and visualization. Jian Cheng, Qiyuan Tian, Weixin Si, Zan Chen, and Shanshan Wang contributed to reviewing and editing. 

\bmsection*{Acknowledgments}
This research was partly supported by the National Natural Science Foundation of China (62222118, U22A2040), Guangdong Provincial Key Laboratory of Artificial Intelligence in Medical Image Analysis and Application (2022B1212010011), Shenzhen Science and Technology Program (RCYX20210706092104034, JCYJ20220531100213029), and Key Laboratory for Magnetic Resonance and Multimodality Imaging of Guangdong Province (2023B1212060052).

\bmsection*{Financial disclosure}

None reported.

\bmsection*{Conflict of interest}

The authors declare no potential conflict of interests.

\bibliography{main}

\end{document}